\documentclass{new_tlp}
\usepackage{mathptmx}
\usepackage{amsmath}
\usepackage{amssymb}
\usepackage{url}
\usepackage{multicol}
\usepackage{enumitem}
\usepackage{wrapfig}

\usepackage{tikz}
\usepgflibrary{shapes.geometric}
\usetikzlibrary{shapes.geometric,shapes.misc,shapes}
\usetikzlibrary{arrows}
\usetikzlibrary{shadows}
\usetikzlibrary{fadings}
\usetikzlibrary{positioning}
\tikzset{vertex/.style={circle,minimum size=6mm,
      very thick,draw=black!50,
      top color=white,bottom color=black!20,inner sep=0pt}}

\newcommand\bcmdtab{\noindent\bgroup\tabcolsep=0pt%
  \begin{tabular}{@{}p{10pc}@{}p{20pc}@{}}}
\newcommand\ecmdtab{\end{tabular}\egroup}

\title[The Power of Non-Ground Rules in ASP]{The Power of Non-Ground Rules in
Answer Set Programming}

\author[M. Bichler, M. Morak and S. Woltran]
{
  MANUEL BICHLER, MICHAEL MORAK and STEFAN WOLTRAN\\
  TU Wien, Vienna, Austria\\
  \email{\{bichler,morak,woltran\}@dbai.tuwien.ac.at}
}

\jdate{March 2003}
\pubyear{2003}
\pagerange{\pageref{firstpage}--\pageref{lastpage}}
\doi{S1471068401001193}

\newenvironment{changemargin}[2]{%
\list{}{\rightmargin#2\leftmargin#1
\parsep=0pt\topsep=0pt\partopsep=0pt}
\item[]}
{\endlist}

\newenvironment{indented}{\begin{changemargin}{1cm}{0cm}}{\end{changemargin}}

\newtheorem{theorem}{Theorem}

\newtheorem{example}[theorem]{Example}

\let\phi\varphi
\let\epsilon\varepsilon

\newcommand{\qedsymbol}{\hfill\ensuremath{\square}}

\newcommand{\calA}{\mathcal{A}}

\newcommand{\calR}{\mathcal{R}}

\newcommand{\calT}{\mathcal{T}}

\newcommand{\NP}{\ensuremath{\textsc{NP}}}
\newcommand{\co}{\ensuremath{\textsc{co}}}

\newcommand{\SIGMA}[2]{\ensuremath{\Sigma_{\textrm{#1}}^{\textrm{#2}}}}

\newcommand{\NEXP}[1][]{\ensuremath{\textsc{N#1ExpTime}}}

\newcommand{\tw}[1]{\mathit{tw}(#1)}

\newcommand{\constants}[1]{\mathbf{#1}}
\newcommand{\variable}[1]{{#1}}
\newcommand{\variables}[1]{{\mathbf{#1}}}

\newcommand{\mods}[1]{\mathit{models}(#1)}
\newcommand{\answersets}[1]{\mathit{AS}(#1)}

\newcommand{\varW}{\variable{W}}
\newcommand{\varX}{\variable{X}}
\newcommand{\varY}{\variable{Y}}
\newcommand{\varZ}{\variable{Z}}
\newcommand{\varsW}{\variables{W}}
\newcommand{\varsX}{\variables{X}}
\newcommand{\varsY}{\variables{Y}}
\newcommand{\varsZ}{\variables{Z}}

\newcommand{\relation}[1]{{\mathit{#1}}}

\newcommand{\fullatom}[2]{{\relation{#1}(#2)}}

\newcommand{\varof}[1]{\mathit{var}(#1)}

\newcommand{\body}[1]{{\mathit{B}(#1)}}
\newcommand{\nbody}[1]{{\mathit{B}^-(#1)}}
\newcommand{\pbody}[1]{{\mathit{B}^+(#1)}}
\newcommand{\head}[1]{{\mathit{H}(#1)}}

\begin{document}

\maketitle

\begin{abstract}
  Answer set programming (ASP) is a well-established logic programming language
  that offers an intuitive, declarative syntax for problem solving. In its
  traditional application, a fixed ASP program for a given problem is designed
  and the actual instance of the problem is fed into the program as a set of
  facts This approach typically results in programs with comparably short and
  simple rules However, as is known from complexity analysis, such an approach
  limits the expressive power of ASP; in fact, an entire NP-check can be encoded
  into a single large rule body of bounded arity that performs both a guess and
  a check within the same rule Here, we propose a novel paradigm for encoding
  hard problems in ASP by making explicit use of large rules which depend on the
  actual instance of the problem. We illustrate how this new encoding paradigm
  can be used, providing examples of problems from the first, second, and even
  third level of the polynomial hierarchy As state-of-the-art solvers are tuned
  towards short rules, rule decomposition is a key technique in the practical
  realization of our approach We also provide some preliminary benchmarks which
  indicate that giving up the convenient way of specifying a fixed program can
  lead to a significant speed-up.

  This paper is under consideration for acceptance in TPLP.
\end{abstract}

\begin{keywords}
  answer set programming, rewriting, non-ground rules, rule decomposition
\end{keywords}

\section{Introduction}\label{sec:introduction}

Answer set programming (ASP)
\cite{
coll:MarekT99,cacm:BrewkaET11,book:GebserKKS12} is a
well-established logic programming paradigm based on the stable model semantics
of logic programs. Its main advantage is an intuitive, declarative language,
and the fact that generally, each answer set of a given logic program describes a valid
answer to the original question. 
Moreover, ASP solvers---see e.g.\
\cite{ai:GebserKS12,lpnmr:AlvianoDFLR13,lpnmr:ElkabaniPS05,datalog:AlvianoFLPPT10}---have
made huge strides in efficiency. 
A logic program usually consists of a set of
logical implications (called rules) and a set of facts. Deciding the consistency
problem, that is, whether a given disjunctive logic program has an answer set,
is \NEXP$^\NP$-complete in the combined complexity, where both the rules and
facts are treated as input, and \SIGMA{P}{2}-complete in the data complexity,
where the set of rules is fixed
; cf.\ \cite{csur:DantsinEGV01}. 

In practice, when problems are modelled using the ASP logic programming
language, the usual goal is to write a fixed program (i.e.\ set of rules) that
solves the general problem. The concrete input is then supplied as a set of
(ground) facts. The answer set solver then takes the fixed program, plus the
ground facts, and computes the answer sets, that is, the solutions to the
original problem, as described earlier. Most ASP programs written in this way
contain only small rules. This is for two main reasons: firstly, current
solvers are much better at evaluating smaller rules, and secondly, large rules
tend not to be as human-readable, and are thus often avoided.  However, 
restricting to fixed programs prohibits
exploiting
the additional expressive power of large rules.  The
following example illustrates
the usage of rules we have in mind.
%
\begin{example}\label{ex:intro}
Consider a graph $G$ over vertices $a,b,c,d$ with edges 
$(a,b)$, 
$(b,c)$, 
$(c,d)$, 
$(a,d)$, and $(b,d)$.
  %
  The problem of three-colorability of 
	this graph
  can be solved by a single, non-ground ASP rule. Let program
  $\Pi$ contain the facts $e(r, g)$, $e(r,b)$, $e(g, b)$, $e(g,r)$, $e(b,r)$, and $e(b, g)$ plus the 
  rule
  $$\bot \leftarrow e(A, B), e(B, C), e(C, D), e(D, A), e(B, D),$$
  where $A, B, C, D$ are variables representing the graph vertices. It is easy
  to verify that the body of this rule is true (and thus, $\Pi$ has no
  answer set) iff each variable can be mapped to one of the constants $r$, $g$,
  and $b$, representing the three colors, such that no two neighboring vertices
  are colored the same. Clearly, a corresponding transformation to the above
  can be applied to any graph, not just to $G$, resulting in a rule whose size
  depends on the graph. \qedsymbol
\end{example}
We postulate that large rules in the spirit of Example~\ref{ex:intro} can be
useful when encoding hard problems into answer set programs, at the expense of
having a single, fixed program solving the problem. When encoding a problem $P$
into ASP in this way, clearly we no longer have a fixed logic program $\Pi$
solving $P$. Instead, a rewriting is used that transforms a specific instance
$I$ of $P$ into a non-ground program $\Pi_I$, with the intention that every
answer set of $\Pi_I$ represents a specific solution to $I$. The rewriting
algorithm is used separately for each concrete instance $I$ of $P$, and thus
can use arbitrarily large, instance-specific rules, that may even encode an
\NP-check (cf.\ Example~\ref{ex:intro}, where a concrete instance of 
3-colorability is rewritten into a corresponding rule).

It is the aim of this paper to propose a general rewriting paradigm that
encodes problems into ASP by making explicit use of large, non-ground rules.
Such rules can, in general, encode \NP-hard checks, even when predicate arities
are bounded (this follows immediately from the \NP-hardness of answering
conjunctive queries over databases \cite{stoc:ChandraM77}). As shown by
\citeN{amai:EiterFFW07}, the combined complexity of solving arbitrary ASP
programs of bounded arity increases to \SIGMA{P}{3}, but drops back to
\SIGMA{P}{2}\ for normal (i.e.\ non-disjunctive) programs. Such programs
therefore fall well within the realm of practically solvable ASP instances, and
we can harness the power of advanced ASP solvers to solve problems above
\SIGMA{P}{2}.

Unfortunately, the use of large rules causes problems for current ASP solvers,
since the input program needs to be grounded first (i.e.\ all the variables in
each rule are replaced by all possible, valid combinations of constants). This
grounding step can be done in polynomial time for fixed ASP encodings, but
generally requires exponential time for rules of arbitrary size. In practice,
the grounding time thus becomes prohibitively large very quickly. However, our
encoding paradigm requires the use of large rules. In order to deal with this
problem, we employ a rule decomposition algorithm that splits large rules into
multiple smaller ones. This approach, based on hypertree decompositions
\cite{jacm:GottlobMS09} of ASP rules, was first proposed by
\citeN{iclp:MorakW12}. We extend this approach to cover the full ASP syntax as
specified by the ASP Standardization Working Group \cite{web:aspcore}. When
such an approach is used, the size of the largest rules is often reduced
drastically, and a traditional grounding/solving approach with existing ASP
solvers becomes feasible for problem encodings according to our paradigm. In
fact, our encodings offers competitive performance when compared to traditional
ASP encodings.

Constructions with large rules, similar to our proposed approach, have been used
in several related areas---usually to establish relevant complexity results.  In
\cite{iandc:GottlobP03}, the complexity of single rule Datalog programs (sirups)
has been investigated. Also, in the world of ontological reasoning and
description logics, rewritings into Datalog (see e.g.\ \cite{kr:GottlobS12})
typically yield rules of large size. \citeN{aaai:EiterFM10} present several
problem encodings with large rules in order to illustrate that current solvers
are not well-equipped to handle them and propose a polynomially space-bounded
solving approach. The grounding bottleneck of state-of-the-art ASP solvers is
widely recognized and several 
methods and systems have been proposed
that either work directly on the non-ground rules via resolution-based methods
\cite{aaai:BonattiPS08}, forward-chaining algorithms \cite{corr:LefevreBSG15},
or lazy grounding approaches where grounding is only performed as needed
\cite{fuin:PaluDPR09,iclp:CatDS12}.

\smallskip \noindent The main results of this paper can be summarized as
follows:
\begin{itemize}[leftmargin=*]
  \item We propose a novel paradigm for problem encoding into ASP that makes
	use of the power of large, non-ground rules. 
	In contrast to the classical ASP approach where a  concrete problem instance $I$ 
	is supplied as ground facts,
	our 
	method
	transforms $I$ into a non-ground program 
	making use of
	arbitrarily large non-ground rules. This allows us to solve problems above
	the class \SIGMA{P}{2}, 
	and exhibits good practical performance
	when compared to classical fixed encodings.

  \item In order to make our paradigm work well in practice, 
we extend the rule decomposition algorithm
	proposed in \cite{iclp:MorakW12} to the full ASP-Core-2 syntax. 
        This allows to feed smaller rules to
        existing ASP solvers, which is a key for good performance. 

  \item We provide a number of case studies that show 
        our proposed encoding
	paradigm at work. 
        We exhibit encodings for solving several
	logic-related problems, including 2- and 3-QBF solving 
        (borrowed from the hardness proofs in \cite{amai:EiterFFW07}), 
	as well as
        an encoding for solving
	ground disjunctive programs via (non-ground) normal programs. 
	Finally, we consider the
	\SIGMA{P}{3}-complete problem of
stable cautious abduction \cite{tcs:EiterGL97}.

  \item We give preliminary benchmark results for 2- and 3-QBF
	solving, comparing the performance of the ASP encodings of the case studies
	to the performance of the classical (fixed) ASP encoding, as well as a
	state-of-the-art QBF solver. 
	Our approach turns out to be surprisingly competitive in practice, 
	even when compared to the dedicated QBF solver.
\end{itemize}

\section{Preliminaries}\label{sec:preliminaries}


\paragraph{Answer Set Programming.} A \emph{ground logic program} (also called
answer set program, or program, for short) is a pair $\Pi = (\calA,\calR)$,
where $\calA$ is a set of propositional (i.e.\ ground) atoms and $\calR$ is a
set of rules of the form:
\begin{equation}\label{eq:rule}
  a_1\vee \cdots \vee a_l \leftarrow a_{l+1}, \ldots, a_m, \neg a_{m+1}, \ldots,
  \neg a_n,
\end{equation}
where $n \geq m \geq l$ and $a_i \in \calA$ for all $1 \leq i \leq n$. A rule
$r \in \calR$ of form~(\ref{eq:rule}) consists of a head $\head{r} = \{
a_1,\ldots,a_l \}$ and a body given by $\pbody{r} = \{a_{l+1},\ldots,a_m \}$
and $\nbody{r} = \{a_{m+1},\ldots,a_n \}$. A rule is called \emph{normal} iff
$l \le 1$. A set $M \subseteq \calA$ is a called a model of $r$ if $\pbody{r}
\subseteq M$ together with $\nbody{r} \cap M = \emptyset$ implies that
$\head{r} \cap M \neq \emptyset$.  We denote the set of models of $r$ by
$\mods{r}$ and the models of a program $\Pi= (\calA,\calR)$ are given by
$\mods{\Pi} = \bigcap_{r \in \calR} \mods{r}$.
The reduct $\Pi^I$ of a program $\Pi$ with respect to a set of atoms $I
\subseteq \calA$ is the program $\Pi^I = \left(\calA,\left\{\head{r} \leftarrow
\pbody{r} \mid r \in \calR, \nbody{r} \cap I = \emptyset)\right\}\right)$.
Following \citeN{iclp:GelfondL88}, $M \subseteq
\calA$ is an \emph{answer set} of a program $\Pi$ 
if $M \in
\mods{\Pi}$ and for no $N \subseteq M$, 
$N \in \mods{\Pi^M}$. 
The set of answer sets of a program $\Pi$ is denoted $\answersets{\Pi}$.
The
\emph{consistency problem} of ASP (decide whether 
for a given program $\Pi$, $\answersets{\Pi}\neq\emptyset$) is $\Sigma^P_2$-complete~\cite{amai:EiterG95}. 

General \emph{non-ground, disjunctive logic programs} differ from ground
programs in that variables may occur in rules. Such rules are
$\forall$-quantified first-order implications of the form $H_1\vee\dots\vee
H_k\leftarrow P_1,\dots,P_n,\neg N_1,\dots,\neg N_m$ where $H_i$, $P_i$ and
$N_i$ are (non-ground) atoms, called head, positive and negative body atoms,
respectively. An \emph{atom} $A$ is of the form $a(\varsX,\constants{c})$ and
consists of a predicate name $a$, as well as a sequence of variables $\varsX$
and a sequence of constants $\constants{c}$ from the domain $\Delta$, with
$|\varsX| + |\constants{c}|$ being the arity of $a$. We will sometimes treat
$\varsX$ and $\constants{c}$ as sets. For zero-arity atoms, we simply write $a$
instead of $a()$. Let $\varof{A}$ denote the set of variables $\varsX$ in atom
$A$. This notation naturally extends to sets of atoms. We will denote variables
by capital letters, constants and predicates by lower-case words. A variable
occurring in the positive body of a rule is said to be \emph{safe}. A rule is
said to be safe if every variable in it is safe. If not stated otherwise, we
always assume that rules are safe. A non-ground rule can be seen as an
abbreviation for all possible instantiations of the variables with domain
elements from $\Delta$. In ASP, this step is usually explicitly performed by a
grounder that transforms a (non-ground) disjunctive logic program into a set
of ground rules of the form given in (\ref{eq:rule}). This process is called
\emph{grounding}. Note that in general, such a ground program can be
exponential in the size of the non-ground program.

\paragraph{Tree Decomposition and Treewidth.} A \emph{tree decomposition} of a
graph $G = (V,E)$ is a pair $\calT = (T, \chi)$, where $T$ is a rooted tree and
$\chi$ is a labelling function over nodes $t$, with $\chi(t) \subseteq V$---we
call $\chi(t)$ the \emph{bag} of $t$---such that the following holds: (i) for
each $v \in V$, there exists a node $t$ in $T$, such that $v \in \chi(t)$; (ii)
for each $\{v,w\} \in E$, there exists a node $t$ in $T$, such that $\{v, w\}
\subseteq \chi(t)$; and (iii) for all nodes $r$, $s$, and $t$ in $T$, such that
$s$ lies on the path from $r$ to $t$, we have $\chi(r) \cap \chi(t) \subseteq
\chi(s)$.  The \emph{width} of a tree decomposition is defined as the
cardinality of its largest bag minus one. The \emph{treewidth} of a graph $G$,
denoted by $\tw{G}$, is the minimum width over all tree decompositions of~$G$.
To decide whether a graph has treewidth at most $k$ is \NP-complete
\cite{siamjadm:ArnborgCP87}. For an arbitrary but fixed $k$ however, this
problem can be solved (and a tree decomposition constructed) in linear time
\cite{siamcomp:Bodlaender96}.
%
Given a non-ground ASP rule $r$, we can
construct its \emph{Gaifman graph} $G = (V, E)$ as follows: Let $V$ be the set
of variables occurring in $r$. Let there be an edge $(X, Y)$ in $E$ iff $X$ and
$Y$ occur together in the head of $r$, or in a body atom of $r$.  We refer to a
tree decomposition of $G$ as the \emph{tree decomposition of rule $r$}.

\section{A New ASP Encoding Paradigm}\label{sec:algorithms}

In this section, we propose a general encoding paradigm allowing to encode hard
problems into non-ground ASP. We first discuss the general method and illustrate
the underlying idea by extending the coloring problem from the introduction.
Secondly, we elaborate 
on the concept of rule decomposition which is a key ingredient towards
practical efficiency.

\subsection{Encoding Hard Problems into ASP using Rule Bodies}

Classically, when solving problems with ASP, the idea is to find a fixed
problem encoding, that is, given a problem $P$, we write a fixed ASP encoding
$\Pi_P$ that solves $P$. An instance $I$ of $P$ is then transformed into a set
of ASP facts (i.e.\ into an input database) $D_I$, and an ASP solver is then
used to solve the ASP program given by $\Pi_P \cup D_I$. The resulting answer
sets usually represent the solutions of the original problem instance $I$.

In contrast, our proposed encoding paradigm facilitates the use of arbitrarily
large non-ground rules. Instead of obtaining a fixed encoding for a problem
$P$, we directly encode the instance $I$ of $P$ into ASP. This requires a
problem-dependent
rewriting algorithm, that takes an instance $I$ of $P$, and generates an ASP
program $\Pi_I$ for $I$. Such a rewriting algorithm may now use, and is in fact
encouraged to use, arbitrarily large rule bodies, that may encode an \NP-hard
check. Note that model-checking for a rule in ASP is already \NP-hard
\cite{iandc:GottlobP03}, which follows from the fact that the body can be
viewed as a conjunctive query, and answering conjunctive queries is known to be
\NP-complete \cite{stoc:ChandraM77}. In the literature, constructions like this
were, to date, mainly used to show theoretic complexity or expressiveness results
by polynomial-time reductions, and not to solve problems in practice. The
following example shows how Example~\ref{ex:intro}, which deals with a concrete
3-colorability instance, can be extended to a rewriting algorithm that deals
with all possible graphs. 

\begin{example}\label{ex:introextended}
  The problem of 3-colorability of an arbitrary input graph $G = (V, E)$ can be
  solved by a single, non-ground ASP rule. Let an ASP program $\Pi$ contain the
  three facts $\mathit{col}(r). \mathit{col}(g). \mathit{col}(b).$ representing
  the tree colors, as well as the three facts $e(r, g). e(g, b). e(b, r).$ and
  their respective inverse, representing valid color pairs. Let rule $r$ be 
  constructed from $G$ as follows:
  $$\bot \leftarrow \bigcup_{v \in V} \{ \mathit{col}(X_v) \} \,\,\, \cup
  \bigcup_{(v, w) \in E} \{ e(X_v, X_w) \},$$
  where for each vertex $v \in V$, a variable $X_v$ represents its color. Note
  that after the first large union, every vertex is guaranteed to be assigned a
  color. The second large union encodes the structure of the graph.  If vertex
  colors can be selected such that the graph structure can be mapped onto the
  valid color pairs, then the body of the rule is true, and the graph is
  3-colorable. Thus, an answer set exists for $\Pi_G$ iff no valid 3-coloring
  exists. \qedsymbol
\end{example}

From the above example, we can see several ingredients needed to encode
problems using our paradigm. Firstly, a (usually fixed) set of ASP facts $D$
contains the domain of our guess (this is done by the $\mathit{col}(\cdot)$
facts in the above example---in particular, we want to guess colors). Further,
the database contains a (usually fixed) set of facts $C$, representing valid,
local combinations of domain elements (this is done by the $e(\cdot, \cdot)$
facts in the above example, which represent the valid color pairs that are
allowed to be next to each other). Finally, we have the large rules that check
for the required properties. Normally, such rules will contain at least one
variable for each item to be guessed. In the above example, we want to guess a
color for each vertex, and thus one variable per vertex is introduced.

In general, such large rules should follow a distinct guess-and-check pattern
for easy readability. The structure of such a large, rewritten rule can thus be
specified as follows:
$$\mathit{head} \leftarrow B_\mathit{guess} \cup B_\mathit{check},$$
where $B_\mathit{guess}$ is a set of body literals encoding the guess, and
$B_\mathit{check}$ is a set of body literals encoding the check. As already
stated, such a rule will contain a variable for each item to be guessed (this
can for example be a variable per vertex to guess colors for the 3-colorability
problem, or a variable per atom to guess a truth assignment for the SAT
problem). The guess-part $B_\mathit{guess}$ will contain atoms that
independently map each variable to one of the domain facts $D$ of the program.
In Example~\ref{ex:introextended}, $B_\mathit{guess}$ contains an atom
$\mathit{col}(X_v)$ for each vertex $v$ of the graph. This represents a guess,
since the only way to satisfy this part of the body is when every variable
$X_v$ is assigned a color. The second part of the body, $B_\mathit{check}$,
contains the actual information about the relation of the guessed variables.
The construction of $B_\mathit{check}$ is highly problem-dependent, but
generally the guessed variables will be mapped to the fixed set $C$ of valid
combinations of domain elements. In Example~\ref{ex:introextended}, this is
done by the atoms $e(X_v, X_w)$ which check that two variables representing two
neighbouring vertices are assigned different colors (since $C$ in this case only
contains valid color pairs for neighboring vertices). Clearly, such large rules
can be combined with other, classical ASP encoding elements. To illustrate this,
the following example further extends Example~\ref{ex:introextended} to a
second-level coloring problem.

\begin{example}\label{ex:introfurtherextended}
  Let $G = (V = V_1 \cup V_2, E)$ be a graph with its vertices partitioned into
  two sets $V_1$ and $V_2$. It can be easily shown that deciding whether there
  exists a three-coloring $C$ for the subgraph of $G$ induced by $V_1$, such
  that there does not exist an extension of $C$ to $V_2$ being a coloring of
  $G$, is \SIGMA{P}{2}-hard. This problem can be solved via a logic program,
  using the same set of facts as Example~\ref{ex:introextended}, and,
  additionally, a set of facts with the predicate $\mathit{vertex}_1$ to
  represent the set $V_1$. This program consists of the following two rules:

  \medskip
  \small
  \qquad$\mathit{c}(X, r) \vee \mathit{c}(X, g) \vee \mathit{c}(X, b) \leftarrow
  \mathit{vertex}_1(X)$, and\qquad $\bot \leftarrow \mathit{edge}(X_1, X_2),
  \mathit{c}(X_1, C), \mathit{c}(X_2, C)$,
  \normalsize
  \medskip

  \noindent
  plus a third rule $r_\mathit{col}$ which is obtained from the rule $r$ of
  Example~\ref{ex:introextended} by adding the body atoms $\{ c(v, X_v) \mid v
  \in V_1 \}$. These additional body atoms require that vertices from $V_1$ are
  assigned the same colors in $r_\mathit{col}$ as in the classical
  guess-and-check part of the program. It can be verified that indeed the above
  logic program has an answer set iff there exists a coloring on $V_1$ that
  cannot be extended to $V_2$ (since the body of $r_\mathit{col}$ is true
  precisely when such an extension exists). \qedsymbol
\end{example}

Note that the above example is a combination of a classical, fixed program
encoding together with large rules according to our paradigm, which illustrates
that the two approaches can be easily combined. 
Moreover, the programs from  
Example~\ref{ex:introfurtherextended}
are head-cycle free (and thus can be easily
transformed into equivalent normal programs via shifting) and their
predicates have a fixed arity; hence, we transform
a $\SIGMA{P}{2}$-complete variant of graph-coloring
into a class of programs for which consistency is of the same
complexity; cf.\
\cite{amai:EiterFFW07}. 
A number of
extended examples how this technique can be applied can be found in
Section~\ref{sec:casestudies}. 

\subsection{Rule Decomposition}\label{sec:ruledecomp}

As we have seen, ASP rules obtained by rewritings as described above may
generally have rule bodies with a large number of atoms (in order to facilitate
an \NP-check). However, state-of-the-art ASP solvers are not well equipped to
handle these large rules. Classically, the program is first converted into a
propositional program by a grounder, which may already lead to problems, since
the number of ground instances of a rule with a large number of body atoms and
variables may be prohibitively large, as the size of the grounding is
exponential in the worst case. 

We thus need a method to make it feasible for large rules to be evaluated with
current grounders and solvers. An approach to split up large, non-ground rules
has been proposed in \cite{iclp:MorakW12}. Generally speaking, this approach
computes the tree decomposition of a rule, and then splits the rule up into
multiple, smaller rules according to this decomposition. \citeN{iclp:MorakW12}
show that, while in the worst case this decomposition may not change the rule at
all, in practice large rules can be split up very well. 
For instance, in 
Example~\ref{ex:introfurtherextended},
for graphs that are sparse (or, more generally speaking, are of low treewidth),
the long rule 
$r_\mathit{col}$ will be amenable for such a decomposition.
Let us briefly recall
the algorithm
from \cite{iclp:MorakW12}. 
 For a given rule $r$, the algorithm works as follows:
\begin{enumerate}[leftmargin=*]
  \item\label{decomp:step1} Compute a tree decomposition $\calT = (T, \chi)$ of
	$r$ with minimal width, with all variables occurring in the head of $r$
	contained in its root node. 

  \item\label{decomp:step2} For each node $n$, let $\mathit{temp}_n$ be a fresh
	predicate, and the same for each variable $\varX$ in $r$ and predicate
	$\mathit{dom}_X$. Let $\varsY_n = \chi(n) \cap \chi(p_n)$, where $p_n$ is
	the parent node of $n$. For the root node $\mathit{root}$, let
	$\mathit{temp}_\mathit{root}$ be the entire head of $r$, and, accordingly,
	$\varsY_\mathit{root} = \varof{\head{r}}$. Now, for a node $n$, generate
	the following rule: 
	\begin{center}
      \begin{tabular}{l l l}
		$\mathit{temp}_n(\varsY_n)$ & $\leftarrow$ & \,\,\; $\{ A
		\in \body{r} \mid \varof{A} \subseteq \chi(n) \}$\\
		& & $\cup \, \{ \mathit{dom}_X(X) \mid A \in \nbody{r}, X \in \varof{A}, \varof{A}
		\subseteq \chi(n) \}$ \\
		& & $\cup \, \{ \mathit{temp}_m(\varsY_m) \mid m \text{ is a child of } n\}.$
      \end{tabular}
    \end{center}

  \item\label{decomp:step3} For each $\varX \in \varof{\nbody{r}}$,
	for which a $\mathit{dom}$ predicate is needed to guarantee safety of a rule
	generated above, pick an atom $A \in \pbody{r}$, such that $\varX \in
	\varof{A}$ and  generate a rule 
	%
	$$\mathit{dom}_X(X) \leftarrow A.$$
\end{enumerate}
Replacing the rule $r$ by the set of generated rules according to the above
algorithm now guarantees that the size of the grounding of $r$ is no longer (at
worst) exponential in the size of the rule, but only exponential in the
treewidth of $r$. Note that the rule decomposition algorithm may also increase
the arity of bounded arity programs, since the arity of temporary predicates
depends on the treewidth. However it can be shown that, since these atoms occur
nowhere except non-negated in the generated rule bodies without internal
recursion, the number of possible answer sets does not increase, and thus the
complexity of bounded arity ASP programs is preserved.

\begin{example}
  Given the rule
  $\fullatom{h}{\varX, \varW} \gets \fullatom{e}{\varX, \varY},
  \fullatom{e}{\varY, \varZ}, \neg \fullatom{e}{\varZ, \varW},
  \fullatom{e}{\varW, \varX}$,
  we may compute the following tree decomposition consisting of two nodes: the
  root node, containing $\{ \fullatom{h}{\varX, \varW}, \fullatom{e}{\varX,
  \varY}, \fullatom{e}{\varW, \varX} \}$, and a child node, containing $\{
  \fullatom{e}{\varY, \varZ}, \neg \fullatom{e}{\varZ, \varW} \}$. Based on this
  decomposition, the rule decomposition algorithm yields the following three
  rules:
  \vspace{-2ex}
  \begin{multicols}{2}
    \small
    \begin{itemize}[leftmargin=*]
      \item $\fullatom{dom_\varW}{\varW} \gets \fullatom{e}{\varW, \varX}$;

      \item $\fullatom{t_1}{\varY, \varW} \gets \fullatom{e}{\varY, \varZ}, \neg
	\fullatom{e}{\varZ, \varW}, \fullatom{dom_\varW}{\varW}$; and

      \item $\fullatom{h}{\varX, \varW} \gets \fullatom{e}{\varX, \varY},
	\fullatom{e}{\varX, \varW}, \fullatom{t_1}{\varY, \varW}$;

      \item[]
    \end{itemize}
  \end{multicols}
  \vspace{-2ex}
  \noindent where $\relation{t_1}$ is a fresh predicate not appearing anywhere
  else.
  \qedsymbol
\end{example}

The above algorithm only deals with rules consisting of simple atoms. However,
in practice, the full answer set syntax contains further, more complicated
atoms; cf.\ \cite{web:aspcore}. In the following, we will explain how the above
algorithm can be extended to the full ASP syntax. Several extensions are
straightforward (e.g.\ rules with upper or lower bounds on the body or head
atoms). We will therefore focus on two extensions that warrant a separate
explanation. Our implementation of the extended rule decomposition algorithm as
described above can be found here: \url{http://dbai.tuwien.ac.at/proj/lpopt}; a
detailed is system description is given in \cite{thesis:Bichler15}. Note that,
since computing an optimal tree decomposition w.r.t.\ width can be \NP-hard, our
implementation employs a heuristic approach for the first step of the above
algorithm; more details can be found in \cite{micai:DermakuGGMMS08}.

\paragraph*{Arithmetics.} An atom in the full language of ASP may be an
\emph{arithmetic atom} of the form $X = \phi$, where $\phi$ is an arithmetic
expression between variables and numbers, connected with the mathematical
connectives $+$, $-$, $\times$, and $\div$. Such an expression impacts the
safety of the generated rules, since $X$ is safe iff all variables in $\phi$ are
safe. In order to deal with this, we change steps \ref{decomp:step2} and
\ref{decomp:step3} of the decomposition algorithm above as follows.

In step \ref{decomp:step2}, a $\mathit{dom}$-predicate is also added for all
variables occurring on the right-hand side of an arithmetic expression. In step
\ref{decomp:step3}, it may now be the case that no atom $A \in \pbody{r}$
exists that contains the variable $X$. In this case, pick the smallest
arithmetic atom of the form $X = \phi$. For each variable $Y_i$ in $\phi$, now
repeat this procedure: Try to pick an atom from $\pbody{r}$ to make $Y_i$ safe.
If no such atom exists, pick the smallest arithmetic atom of the form $Y_i =
\psi$ not already selected. This procedure necessarily terminates, and selects
a set of atoms $\mathbf{A}$ from $r$. We generate the rule $\mathit{dom}_X(X)
\leftarrow \mathbf{A}$. It is easy to see that this rule is safe, and describes
the possible domain of variable $X$, as required.

\paragraph*{Aggregates.} An aggregate is an atom $A$ of the form $\#agg\{ \varsX
\,:\, \Psi(\varsX, \varsY, \varsZ) \}$, where $\Psi$ is a conjunction of
non-ground literals over the variables in $\varsX$, $\varsY$ and $\varsZ$.
Intuitively, the variables in $\varsX$ are used for aggregation. $\varsY$ are
those variables that also occur outside the aggregate, and $\varsZ$ are those
variables appearing within the aggregate only. Such an aggregate atom $A$ is
rewritten as follows.

First, replace $A$ by an aggregate atom $A'$ that only preserves those atoms
within $A$ that have connections to the rest of the rule $r$ containing $A$.
To this end, let $A'$ be of the form $ \#agg\{ \varsX \, : \, \Psi'(\varsX,
\varsY, \varsW), \mathit{temp}_A(\varsX, \varsY, \varsW) \}$, where
$\Psi'(\varsX, \varsY, \varsW) = \{ B \in \Psi \mid \varof{B} \cap \varsX \neq
\emptyset \vee \varof{B} \cap \varsY \neq \emptyset \}$, $\mathit{temp}_A$ is a
fresh predicate, and $\varsW = (\varsZ \cap \varof{\Psi'})$, that is, all those
variables from $\varsZ$ occurring in $\Psi'$. Then, create a temporary rule
$r_A = \mathit{temp}_A(\varsX, \varsY, \varsW) \leftarrow (\Psi \setminus \Psi'
\cup \Psi_\mathit{dom})$, that is, $r_A$ contains all the atoms of $A$ that
have no connection to the rest of $r$, plus the set $\Psi_\mathit{dom}$
containing $\mathit{dom}_X$-predicates for each otherwise unsafe variable $X$
in $r_A$. Finally, recursively execute the rule decomposition algorithm on
$r_A$. Since aggregates may be arbitrarily large, this allows us to decompose
large aggregates to a smaller aggregate and a set of smaller rules.


\section{Case Studies}\label{sec:casestudies}

In this section we exhibit a number of interesting problems that can be solved
using our proposed encoding paradigm. The problems include evaluating 2-QBF and
3-QBF formulas, as well as solving disjunctive ground ASP itself. The latter is
then extended to an encoding for abduction.

\subsection{Rewriting QSAT into Logic Programs}
\label{sec:qsatrewriting}

Before introducing the encodings according to our paradigm, we give a classic,
fixed-program encoding for 2-QBFs in the spirit of the encoding used in the
$\SIGMA{P}{2}$-completeness proof from \cite{amai:EiterG95}. This encoding will
be used for comparison in the benchmarks in Section~\ref{sec:evaluation}.
%
%
Let $\Phi$ be a fully quantified 2-QBF of the form
\begin{equation}\label{eq:1}
\Phi = \forall x_1, \ldots, x_m \, \exists y_1, \ldots, y_n \, \, \, ( c_1
\land \ldots \land c_k) ,
\end{equation}
where clauses are of the form $c_i = L_{i,1} \vee L_{i,2} \vee L_{i,3}$ with
literals $L_{i,j}$ over $\{x_1, \ldots, x_m , y_1, \ldots, y_n\}$.  
We construct $\Pi^\Phi$ which consists of the following fixed set of rules:
\vspace{-2ex}
\begin{multicols}{2}
  \small
  \begin{itemize}
	\item $\mathit{ass}(X, 1) \vee \mathit{ass}(X, 0) \leftarrow \mathit{var}(X)$;
	  
	\item $\mathit{ass}(X, 0) \leftarrow \mathit{sat}, \mathit{exists}(X)$; 
	\item $\mathit{ass}(X, 1) \leftarrow \mathit{sat}, \mathit{exists}(X)$; 
	  
	\item $\mathit{sat} \leftarrow \bigcup_{i \leq 3} \{ \mathit{pos}_i(C, X_i, A_i), \mathit{ass}(X_i, 1 - A_i) \}$;

	\item $\bot \leftarrow \neg \mathit{sat}$;
	\item[]
  \end{itemize}
\end{multicols}
\vspace{-2ex}
\noindent
and facts derived from $\Phi$ as follows: a fact $\mathit{var}(x_i)$ for each
variable $x_i$, a fact $\mathit{exists}(x_i)$ if $x_i$ is existentially
quantified, and a fact $\mathit{pos}_\ell(c_i, x_j, a_\ell)$ for each occurrence
of a variable $x_j$ in a clause $c_i$ at position $\ell \in \{ 1, 2, 3 \}$,
where $a_\ell$ is $0$ if $x_j$ appears negated in $c_i$; otherwise $a_\ell$ is $1$.
It can be checked that $\Pi^\Phi$ has an answer set iff $\Phi$ is
unsatisfiable.

A different reduction algorithm for QBFs of the form (\ref{eq:1}) has been proposed in
\cite{amai:EiterFFW07}, which in fact makes use of the paradigm proposed in
Section~\ref{sec:algorithms}. Let $\Phi$ be a QBF as before. For ease of
notation, let $X(c_i) = \{ L \mid L \mbox{ is a literal of form $x_j$ or $\neg x_j$ in } c_i \}$, and let
$Y(c_i)$ be the ordered tuple of the literals in $c_i$ over the variables
$\{y_1, \ldots, y_n\}$.  Further, let $\eta(c_i)$ be the tuple obtained by
replacing each literal of $Y(c_i)$ with a corresponding (ASP) variable, ignoring
negation, and let $\overline{c_i}$ denote the tuple resulting from $Y(c_i)$ by
replacing each positive literal in $Y(c_i)$ by $0$ and each negative literal by
$1$.  For example, for $c=\lnot y_1 \lor x_2 \lor y_3$, we have $Y(c)=(\neg
Y_1,Y_3)$, $\eta(c) = (Y_1, Y_3)$, and $\overline{c}=(1,0)$.
Let $\Pi^\Phi$ contain the following rules:
\begin{itemize}\small
  \item $t(x_i) \vee f(x_i)$, for each $i \in \{1,\ldots,m\}$;

  \item $c_i(\mathbf{t}) \leftarrow t(x_j)$, for each $i \in \{1,\ldots,k\}$, tuple
	$\mathbf{t} \in \{0, 1\}^{|Y(c_i)|}$, and positive literal $x_j \in X(c_i)$;
  \item $c_i(\mathbf{t}) \leftarrow f(x_j)$, for each $i \in \{1,\ldots,k\}$, tuple
	$\mathbf{t} \in \{0, 1\}^{|Y(c_i)|}$, and negative literal $\neg x_j \in X(c_i)$;

  \item $c_i(\mathbf{t})$, for each $i \in \{1,\ldots,k\}$ and tuple $\mathbf{t} \in \{0,
	1\}^{|Y(c_i)|} \setminus \{ \overline{c_i} \}$; and 

  \item $\bot \leftarrow c_1(\eta(c_1)), \dots, c_k(\eta(c_k))$.
\end{itemize}
%
Note that predicate arities in the above construction are bounded by the
constant $3$ and the last rule is the only rule containing variables, and has
a body with size linear in the size of $\Phi$.  Moreover, the program is
head-cycle free and thus can equivalently be given as normal program.  From
\cite{amai:EiterFFW07}, we again have that $\Pi^\Phi$ has an answer set iff
$\Phi$ is unsatisfiable.

In \cite{amai:EiterFFW07} the above approach is then extended to solve
third-level QBFs, effectively reducing a $\SIGMA{P}{3}$-complete problem to
disjunctive ASP. To this end, $\Phi$ be a 3-QBF as follows: $$\Phi =
\exists x_1, \ldots, x_l \forall x_{l+1}, \ldots x_m \exists y_1 \ldots y_n (c_1
\land \ldots \land c_k).$$ A program $\Pi^\Phi$ can be constructed from $\Phi$
as before (by treating all $x_i$ variables as if they were universally
quantified). By replacing the $\bot$ symbol with an atom $\mathit{sat}$, adding
the rules $\bot \leftarrow \neg \mathit{sat}$ and 
$p \gets \mathit{sat}$ for each atom
$p \in \{ t(x_i), f(x_i) \mid l < i \leq m \} \cup
       \{ c_i(t) \mid 0 < i \leq k, t \in \{ 0, 1 \}^{ |Y(c_i)| } \}$,
the 2-QBF construction
can be extended to 3-QBF formulas $\Phi$, such that $\Pi^\Phi$ has an answer set
iff $\Phi$ is valid.

\subsection{Rewriting Disjunctive ASP into Normal Logic Programs}
\label{sec:rewritingasp}

In this subsection, we use our method to rewrite disjunctive (ground) ASP
programs to non-ground \emph{normal} ASP programs. In order words, we ``shift''
the additional complexity caused by disjunctive rule heads into non-ground rule
bodies. As in the previous subsection, our encodings require predicates of fixed
arity only, thus we map a \SIGMA{P}{2}\ problem to another \SIGMA{P}{2}\ problem. 

Let $\Pi = (\calA, \calR)$ be a ground, disjunctive logic program. 
We construct a non-ground, normal logic program $\hat{\Pi}$ of bounded arity,
such that every answer set of $\hat{\Pi}$ witnesses the existence of a
corresponding answer set of $\Pi$. 
%
First,
$\hat{\Pi}$ contains the following facts: 
\begin{multicols}{2}
  \small
  \begin{itemize}
	\item 
$\{ \mathit{atom}(a) \mid a\in\calA\}$,
	\item 
$\{ \mathit{rule}(r) \mid r\in\calR\}$,
\item 
$\{ \mathit{leq}(0, 0), \mathit{leq}(0, 1), \mathit{leq}(1, 1)\}$,
\item 
$\{ \mathit{or}(0, 0, 0), \mathit{or}(0, 1, 1), \mathit{or}(1, 0, 1), \mathit{or}(1, 1, 1)\}$,
	\item 
$\{ \mathit{head}(r, a) \mid r\in\calR, a\in\head{r}\}$,
	\item 
$\{ \mathit{pos}(r, a) \mid r\in\calR, a\in\pbody{r}\}$, and
	\item 
$\{ \mathit{neg}(r, a) \mid r\in\calR, a\in\nbody{r}\}$.
	\item[]
  \end{itemize}
\end{multicols}
%
\noindent Predicates $\mathit{atom}$, $\mathit{rule}$, 
$\mathit{head}$, $\mathit{pos}$, and $\mathit{neg}$ describe the disjunctive ground program 
to be evaluated.
Predicate 
$\mathit{leq}$ represents the less-or-equal relation;
predicate 
$\mathit{or}$
is used to
encode logical disjunction.  
The role of the latter two predicates will be clarified below.
%
Next, we give a fixed
set of six rules that guesses an assignment on the atoms of $\Pi$, and then
checks that this assignment is a classical model of $\Pi$:
\begin{multicols}{2}
  \small
  \begin{itemize}
	\item $\mathit{assign}(A, 1) \leftarrow \mathit{atom}(A), \neg
	  \mathit{assign}(A, 0)$;

	\item $\mathit{assign}(A, 0) \leftarrow \mathit{atom}(A), \neg
	  \mathit{assign}(A, 1)$;

	\item $\mathit{sat}(R) \leftarrow \mathit{head}(R, A), \mathit{assign}(A, 1)$;

	\item $\mathit{sat}(R) \leftarrow \mathit{pos}(R, A), \mathit{assign}(A, 0)$;

	\item $\mathit{sat}(R) \leftarrow \mathit{neg}(R, A), \mathit{assign}(A, 1)$; and

	\item $\bot \leftarrow \mathit{rule}(R), \neg \mathit{sat}(R)$.
  \end{itemize}
\end{multicols}
\noindent Finally, we need to check the subset-minimality of the guessed
assignment $M$ w.r.t.\ the reduct $\Pi^M$. This is done with one large rule
$r_\mathit{reduct}$, encoding the guess of a subset $M' \subset M$, as well as
the check if this subset $M'$ satisfies all the rules in $\Pi^M$.  In that
case, the guessed assignment cannot be an answer set and the constraint
$r_\mathit{reduct}$ fires.

Let the last rule $r_\mathit{reduct}$ be of the following form, where
$\mathbf{X}$ and $\mathbf{Y}$ contain a variable $X_a$ and $Y_a$ for each atom
$a \in \calA$, representing the atom's truth assignment in $M$ and $M'$,
respectively:
$$r_\mathit{reduct} = \bot \leftarrow B_\mathit{subset}(\mathbf{X}, \mathbf{Y})
\cup B_\mathit{neq}(\mathbf{X}, \mathbf{Y}) \cup B_\mathit{model}(\mathbf{X},
\mathbf{Y}),$$
where the literals in $B_\mathit{subset}$ encode the fact that $M'$ (represented
by $\mathbf{Y}$) is a subset of $M$ (represented by $\mathbf{X}$);
$B_\mathit{neq}$ encodes the fact that this subset relation is proper; and
$B_\mathit{model}$ encodes the model check of $M'$ against the reduct $\Pi^M$.
Since ASP does not allow disjunction in rule bodies, we will use the
$\mathit{or}(\cdot,\cdot,\cdot)$ atoms to simulate disjunction.
$B_\mathit{subset}$ first guesses a truth value for each variable in $\varsY$
that is smaller or equal to the corresponding variable in $\varsX$ as follows:
%
$$B_\mathit{subset}(\mathbf{X}, \mathbf{Y}) \equiv \bigcup_{a \in \calA} \{ \mathit{assign}(a, X_a),
\mathit{leq}(Y_a, X_a) \}.$$
Then, in $B_\mathit{neq}$, we use our $\mathit{or}$ atoms as disjunction to
check that at least one truth value in $\varsY$ is different from the one in
$\varsX$, which, in combination with $B_\mathit{subset}$, guarantees that the
true atoms in $\varsY$ are a proper subset of the ones in $\varsX$:
$$B_\mathit{neq}(\mathbf{X}, \mathbf{Y}) \equiv \{ N_0 = 0 \} \cup \bigcup_{a \in \calA} \{
\mathit{or}(N_i, X_a - Y_a, N_{i + 1}) \} \cup \{ N_{|\calA|} = 1 \}.$$
Finally, we check that the truth assignment stored in $\varsY$ actually
represents a model of the reduct. To this end, $B_\mathit{model}$ checks, by
again making use of the $\mathit{or}$ atoms to encode disjunction, that in each
rule there exists an atom that makes the rule true. We have:
$$
  B_\mathit{model}(\mathbf{X}, \mathbf{Y}) \equiv \bigcup_{r \in \calR} \big( \{ R^r_0 = 0 \} \cup
  \bigcup_{a \in \head{r}} \{ \mathit{or}(R^r_i, Y_a, R^r_{i + 1}) \} \,\,\,\,\, \cup
  \bigcup_{a \in \pbody{r}} \{ \mathit{or}(R^r_j, 1 - Y_a, R^r_{j + 1}) \} \,
  \cup 
$$
$$
  %
\quad
\quad
\quad
\quad
	\bigcup_{a \in \nbody{r}} \{ \mathit{or}(R^r_k, X_a, R^r_{k + 1})
  \} \,\,\,\,\, \cup \,\,\,\,\, \{ R^r_{|r|} = 1 \} \big),
$$
where for a rule $r \in
\calR$, $|r| = |\head{r}| + |\pbody{r}| + |\nbody{r}|$. 
In the above, variables $N$ and $R$ are used to ``glue'' together the disjunctions. 
For instance, in $B_\mathit{neq}$ we have that $N_i$ switches from $0$ 
to $1$ for $N_{i+1}$ as soon as there is at least one atom assigned true in $M$ but false in $M'$, i.e.\ when $M'\subset M$.
For the sake of readability, we do not explicitly specify how $i$, $j$ and $k$
are determined, but we assume that these index numbers increment by one for each
element added to the respective set. Clearly, this could easily be formalized by
assuming an order over all the atoms and rules. 

Note that in the above construction, we have a fixed, non-disjunctive logic
program to check for (classical) satisfiability. Then, according to the encoding
paradigm proposed in Section~\ref{sec:algorithms}, we use a rule with a large
body to encode the \co\NP-check for stability (i.e.\ checking that a given
classical model is minimal w.r.t.\ its reduct). This construction actually works
for any ground, disjunctive answer set program, and thus yields a kind of
``meta-solver'' for ASP: we can use the rewriting above, and an ASP solver, to
solve ASP itself. Another interesting observation is that, if we run a grounder
on this program, we obtain a ground, normal logic program $\Pi'$ (i.e.\ without
disjunction) that solves our original ground logic program $\Pi$ (which may
contain disjunction). Thus, we have a rewriting algorithm that eliminates
disjunction from ground logic programs, at the cost of an exponential blowup in
general (since the grounding of $r_\mathit{reduct}$ may be exponential).

The idea of creating such a meta-solver using reification techniques is not new;
see e.g.\ \cite{tplp:GebserKS11} and the references therein.  However, most of
these approaches ``stay'' within the same class of programs, i.e.\ the
meta-program to solve a disjunctive logic program is itself a disjunctive logic
program.  Some of the meta-programming approaches in fact take the other
direction than we do here.  For example, in \cite{tplp:EiterP06} a meta-solver
is used to equip a problem formulated as a non-disjunctive program with an
additional \co\NP-test, yielding a disjunctive program as a result. Similar
approaches are used to handle optimization statements
\cite{tplp:GebserKS11,aaai:BrewkaDRS15}.  As we have emphasized before, our aim
here is the opposite direction, that is, translating a ground disjunctive
program into a (non-ground) program from an easier class\footnote{The
\emph{GnT} system follows an alternative idea to evaluate disjunctive logic
programs \cite{tocl:JanhunenNSSY06}: it facilitates a generate-and-test method
based on disjunction-free programs, thus relying on two calls of an ASP-solver; in the worst-case an exponential input for
the second program needs to be generated.}.  For existing ASP solvers, solving
performance drops significantly when (full) disjunction comes into play. It will
be thus interesting to see whether such a rewriting improves performance. 

\subsection{Rewriting Stable Cautious Inference into Disjunctive Logic Programs}

Finally, we provide an encoding for the \SIGMA{P}{3}-complete
problem of \emph{stable cautious inference} 
\cite{tcs:EiterGL97}, 
defined as follows: Given a
tuple $\langle \Pi, H, M\rangle$, where $\Pi = (\calA, \calR)$ is a ground
(disjunctive) logic program, and $H \subseteq \calA$ and $M \subseteq \calA$ are
sets of (ground) atoms called \emph{hypotheses} and \emph{manifestations},
respectively: decide whether there exist a subset $E \subseteq H$, such that for
all answer sets $S$ of $\Pi \cup E$ it holds that $M \subseteq S$.
Note that the original definition also requires that $\answersets{\Pi \cup E}\neq\emptyset$ 
However, inspecting the proof in \cite{tcs:EiterGL97} shows that even without
this condition, \SIGMA{P}{3}-hardness is preserved. Thus, to simplify our
construction, we will omit it.

The encoding of this program will partly be an adaptation of the encoding 
from Section~\ref{sec:rewritingasp}; in particular, we will reuse the idea of
$r_\mathit{reduct}$.
The aim is to find an assignment to the hypotheses, such that 
for any assignment of the remaining atoms \emph{not} containing all
manifestations it is the case that such a joint assignment is \emph{not}
an answer set. We do so by employing saturation. This also forces us 
to encode the test for classical satisfaction of the rules in a different
way than in Section~\ref{sec:rewritingasp}, i.e.\ we have to saturate 
as soon as one rule is not satisfied.

To this end, let $\langle \Pi, H, M \rangle$ be an instance of the stable
cautious inference problem. We create a program $\widehat{\langle \Pi,H,M
\rangle}$ that has an answer set iff the tuple $\langle \Pi, H, M \rangle$
represents a valid instance of stable cautious inference. 
%
We need only the following facts: 
for each atom $a$ in $\Pi$, $\mathit{atom}(a)$; 
and for each $a\in H$,
$\mathit{hyp}(a)$. 
%
The fixed set of rules contains the following:
\begin{multicols}{2}
  \small
  \begin{itemize}
	\item $\mathit{assign}(A, 1) \vee \mathit{assign}(A, 0) \leftarrow
	  \mathit{atom}(A)$;






	\item $\mathit{assign}(A, 1) \leftarrow \mathit{sat}, \mathit{atom}(A), \neg
	  \mathit{hyp}(A)$;

	\item $\mathit{assign}(A, 0) \leftarrow \mathit{sat}, \mathit{atom}(A), \neg
	  \mathit{hyp}(A)$; and

	\item $\bot \leftarrow \neg \mathit{sat}$.
  \end{itemize}
\end{multicols}
\noindent
Then, we have several rules for saturation. First, we saturate if all manifestations are in the guess:
$$
\mathit{sat} \leftarrow  \bigcup_{a\in M} \{ \mathit{assign}(a, 1)  \}.
$$
Second, we saturate if the assignment is not a model of the 
program $\Pi$ (and thus not of $\Pi\cup E$). To this end, for every rule $r \in \calR$ we add 
$$
\mathit{sat} \leftarrow  
\bigcup_{a\in H(r)} \{ \mathit{assign}(a, 0)  \} \cup
\bigcup_{b\in B^+(r)} \{ \mathit{assign}(b, 1)  \} \cup
\bigcup_{c\in B^-(r)} \{ \mathit{assign}(c, 0)  \}.
$$
Finally, we again need the reduct check. Let the last rule again be
$r_\mathit{reduct}$, defined in the same way as in the construction in
Section~\ref{sec:rewritingasp}:
$$r_\mathit{reduct} = \mathit{sat} \leftarrow B_\mathit{subset}(\mathbf{X},
\mathbf{Y}) \cup B_\mathit{neq}(\mathbf{X}, \mathbf{Y}) \cup
B_\mathit{model}(\mathbf{X}, \mathbf{Y}).$$
In $r_\mathit{reduct}$, the last two sets remain the same as in
Section~\ref{sec:rewritingasp}. The first set, $B_\mathit{subset}$, however,
needs to reflect that the guessed hypotheses $E$ are indeed facts 
in the program $\Pi\cup E$
(i.e.\ if they
are true in a model candidate $M$ represented by the variables $\mathbf{X}$,
they must also be true in the reduct model candidate $M' \subset M$
represented by the variables $\mathbf{Y}$) . Thus, $B_\mathit{subset}$ is
changed as follows:
$$B_\mathit{subset} \equiv \bigcup_{a \in \calA \setminus H} \{
\mathit{assign}(a, X_a), \mathit{leq}(Y_a, X_a) \} \,\,\, \cup \bigcup_{a \in H}
\{ \mathit{assign}(a, X_a), X_a = Y_a \}$$
This completes the construction. 
Now, if a subset $E$ of hypotheses can be found, such that saturation 
is not applied (observe that we saturate over all non-hypotheses atoms),
we know that there is an assignment extending $E$ to all atoms
that (i) does not contain all manifestations, 
(ii) is a model of $\Pi \cup E$, and 
(iii) there is no subset of that assignment 
that is a model of the reduct. Hence we found an answer of $\Pi\cup E$ that does not contain all manifestations.
In that case, due to saturation, and the rule
	$\bot \leftarrow \neg \mathit{sat}$,
this particular guess for $E$ will not yield an answer set 
of the encoding.
Thus
an answer set $S$ of the
rewritten program $\widehat{\langle \Pi, H, M \rangle}$ represents a valid
selection of hypotheses $E$ from $H$, such that in every answer set of $\Pi
\cup E$ we find all atoms from $M$. 

\section{Experimental Evaluation}\label{sec:evaluation}

In this section, we give a preliminary experimental evaluation of how our
proposed rewritings perform when compared to current encodings and
state-of-the-art problem-specific solvers.  We have implemented the QBF
rewriting algorithms of Section~\ref{sec:qsatrewriting} 
and
integrated these rewritings with the extended rule decomposition tool,
\emph{lpopt}\footnote{The updated \emph{lpopt} tool is available at
\url{http://www.dbai.tuwien.ac.at/proj/lpopt}}, as described in
Section~\ref{sec:ruledecomp}.

We used publicly available 2-QBF ($\forall\exists$) competition
instances\footnote{See \url{http://www.qbflib.org/TS2010/2QBF.tar.gz}.} and
instances from the Eval-2012 data set\footnote{Available at
\url{http://qbf.satisfiability.org/gallery/eval2012r2.tgz}.} of the latest
QBF competition for 3-QBF benchmarks. All benchmark inputs can be found
online\footnote{\url{http://dbai.tuwien.ac.at/proj/lpopt/benchmarks.tgz}}.
In the following, we compare the performance of (1) the classic encodings for
2-QBFs as described in Section~\ref{sec:qsatrewriting}; (2) the 
new encoding according to our paradigm as described in
Section~\ref{sec:qsatrewriting}, where the long rules are decomposed with 
\emph{lpopt}; and
(3) a dedicated QBF solver. For (1) and (2), we first perform \emph{DepQBF} preprocessing \cite{lpar:LonsingBBES15} and employ the well-known ASP solver
\emph{clingo}\footnote{\url{http://potassco.sourceforge.net/}} on the preprocessed instances. For (3), we
employ the state-of-the-art QBF solver
\emph{DepQBF}\footnote{\url{http://lonsing.github.io/depqbf/}}. All tools were
used in their most recent version. 3-QBF-benchmarks were run on a 16-core AMD
Opteron machine with 2.1GHz, 224 GB of RAM, running Debian Linux.
2-QBF-benchmarks were run on an 8-core Intel Xeon machine with 2.33GHz and 48 GB
of RAM. A global timeout was set at 600 seconds for 2-QBF, and 900 seconds for
3-QBF instances. All times measured are cumulative, that is, we sum up the times
used for \emph{DepQBF} preprocessing, rule decomposition, grounding, and solving. We measured the CPU time,
thus, time wasted for I/O operations is not included in our measurements.

\begin{figure}
  \begin{multicols}{2}
	\includegraphics[width=.9\linewidth]{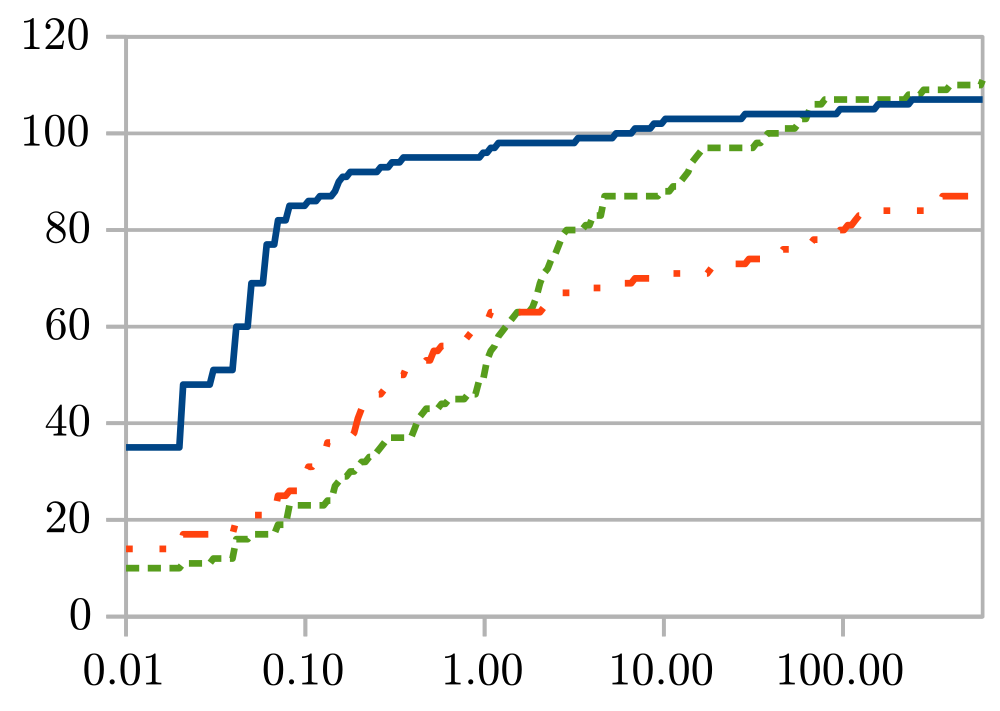}

	\vspace{-1.7ex}
	\begin{center} (a) \end{center}

	\includegraphics[width=.9\linewidth]{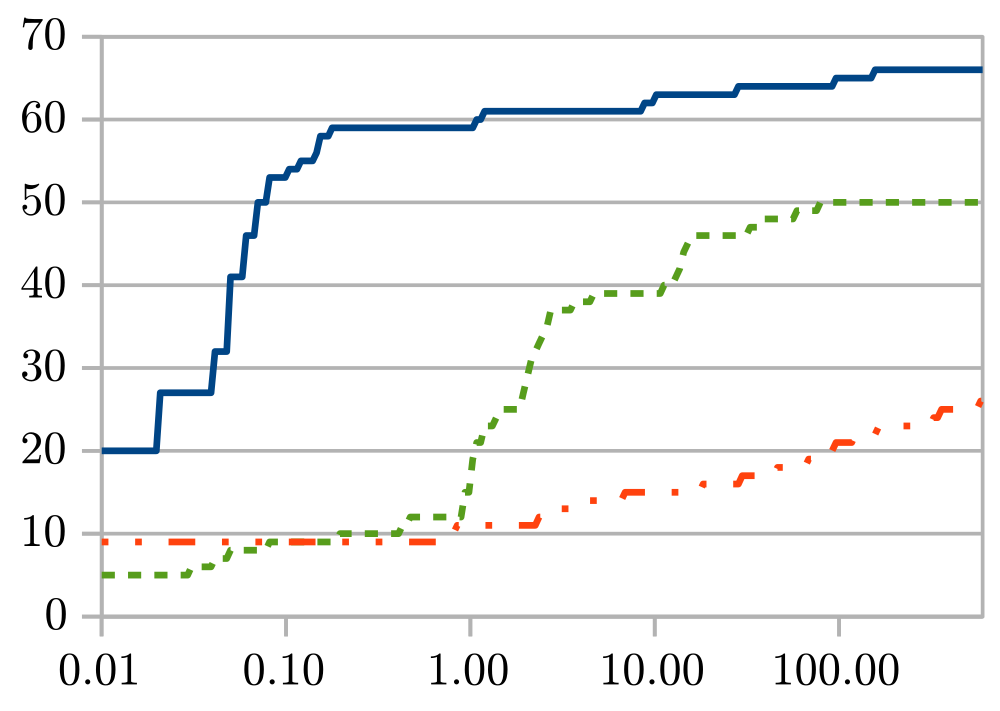}

	\vspace{-1.7ex}
	\begin{center} (b) \end{center}
  \end{multicols}
  \vspace{-2ex}
  \caption{Solved 2-QBF instances on the y-axis vs. time on the x-axis. Classic
	encodings in dotted-dashed red, encodings according to our paradigm in dashed green, and
	\emph{DepQBF} in solid blue. (a) shows the overall benchmark results, (b) only the
	results for the ``stmt'' problem.}
  \vspace{-2ex}
  \label{fig:2qbf}
\end{figure}

\paragraph*{2-QBF Results.} As can be seen in Figure~\ref{fig:2qbf}(a), results
for 2-QBF solving are very encouraging. In particular, for a runtime of about 2
seconds, we match the performance of the classical encoding, and for longer
runtimes surpass it significantly overall. Where the classical encoding could
only solve 88 instances in total, with our encoding according to the paradigm
proposed in Section~\ref{sec:algorithms}, \emph{clingo} was able to solve 111
instances successfully within the given timeout window. Surprisingly, this was
even better than \emph{DepQBF}, a solver built specifically to evaluate QBF
formulas. It was only able to solve 107 instances out of our benchmark set of
200 instances.

The largest single instance set within the benchmark set was the ``stmt''
problem. The benchmark results are shown in Figure~\ref{fig:2qbf}(b). While
here, \emph{DepQBF} performed better than the ASP encodings, this still shows
that our encoding can indeed outperform the classical encoding by a large
margin, solving 51 instances compared to 26. Note that we do not claim to beat
\emph{DepQBF} in general in terms of performance when solving QBF problems.
However, the benchmarks clearly show that our encoding paradigm can be by far
superior to the classical, fixed ASP encoding.

\vspace{-1.5ex}
\paragraph*{3-QBF Results.} In the case of solving third-level problems,
results are sparse. For the 151 3-QBF instances, \emph{DepQBF} was able to
solve 47, while our approach was able to solve 18. However, out of these 18
instances, there were ten that our 3-QBF encoding was able solve, while
\emph{DepQBF} was unable to solve them within the time limit. This again
indicates the viability of our encoding paradigm, using the power of readily
available, state-of-the-art ASP solvers in order to solve problems on the third
level of the polynomial hierarchy.

\smallskip
Let us conclude with a remark on the treewidth of the large rules in our
encodings. Generally, current grounders tend to time out when a rule is too
large. Thus, when encoding problems with large rules according to our paradigm,
performance is best when the resulting rule has low treewidth, since then our
rule decomposition tool can split the rule up into very small rules that are
easy to process for the grounder. When the treewidth of the rule is high
however, the decomposed rules will still be very large, in which case the
grounding bottleneck may still be a dominating factor. Rule treewidth in our
instances ranged from 3 to over 300, and grounding sizes (and timeouts) varied
accordingly, leading to expected grounder timeouts on instances with large,
high-treewidth rules. However, we note that this is a property of the original
input instance, and not of our rule decomposition approach which, in the worst
case, will not change the rules in the program at all, and can thus not make
things worse. Finally, since \emph{lpopt} makes use of heuristics to compute
tree decompositions (see \cite{micai:DermakuGGMMS08} for details of the
heuristics used), some variability in decomposition quality is expected. This
can cause variations in grounding time and size for different decompositions of
the same rule. However, these generally stayed within the range of less than
10\%.

\section{Conclusion}\label{sec:conclusions}

In this paper, we laid out a novel approach to encode problems into ASP. These
rewritings make heavy use of large rules, which can be used to encode
\NP-checks. For disjunctive ASP, if the predicate arity is bounded, it becomes
possible to solve problems of up to \SIGMA{P}{3}-hardness with this approach,
since the consistency problem for disjunctive ASP with bounded predicate arity
is itself \SIGMA{P}{3}-complete \cite{amai:EiterFFW07}. Because existing
grounders are slow to ground large rules, we then present an extended rule
decomposition algorithm to make our encodings solvable in practice. Finally, we
provided several examples on how our approach can be used to solve problems on
the second and third level of the polynomial hierarchy, and show that these
encodings actually perform well in practice.

\paragraph*{Acknowledgments.} This work was funded 
by the Austrian Science Fund (FWF): Y698, P25607.

\bibliographystyle{acmtrans}
\bibliography{iclp2016}

\end{document}